\documentclass[preprint,12pt]{elsarticle}

\usepackage{amsmath,amssymb,amsfonts}
\usepackage{algorithmic}
\usepackage{graphicx}
\usepackage{textcomp}
\usepackage{xcolor}
\usepackage[colorlinks=true]{hyperref}
\usepackage{tablefootnote}
\def\BibTeX{{\rm B\kern-.05em{\sc i\kern-.025em b}\kern-.08em
    T\kern-.1667em\lower.7ex\hbox{E}\kern-.125emX}}

\usepackage{spverbatim}
\usepackage{multirow}
\usepackage{booktabs}
\usepackage[caption=false]{subfig}

\newcommand{\coderepositoryurl}{https://github.com/AlgTUDelft/ExpensiveOptimBenchmark}

\newcommand\bname{EXPObench} 

\DeclareMathOperator*{\argmax}{\arg\!\max}

\newcommand{\editA}[1]{#1}
\newcommand{\editB}[1]{#1}

\journal{Applied Soft Computing Journal}

\begin{document}

\begin{frontmatter}

\title{EXPObench: Benchmarking Surrogate-based Optimisation Algorithms on Expensive Black-box Functions
}

\author[tue]{Laurens Bliek}
\author[tud]{Arthur Guijt}
\author[tud]{Rickard Karlsson}
\author[tud]{Sicco Verwer}
\author[tud]{Mathijs de Weerdt}

\address[tue]{{School of Industrial Engineering, Eindhoven University of Technology},
            {\ PO Box 513, 5600 MB}, 
            {Eindhoven}, 
            {the Netherlands}}
            
\address[tud]{{Faculty of Electrical Engineering, Mathematics and Computer Science, Delft University of Technology},
            {\ PO Box 5
2600 AA},
            {Delft}, 
            {the Netherlands}}




 \begin{abstract}
 Surrogate algorithms such as Bayesian optimisation are especially designed for black-box optimisation problems with expensive objectives, such as hyperparameter tuning or simulation-based optimisation. In the literature, these algorithms are usually evaluated with synthetic benchmarks which are well established but have no expensive objective, and only on one or two real-life applications which vary wildly between papers. There is a clear lack of standardisation when it comes to benchmarking surrogate algorithms on real-life, expensive, black-box objective functions. This makes it very difficult to draw conclusions on the effect of algorithmic contributions
 \editA{and  to give substantial advice on which method to use when.}
A new benchmark library, 
\bname, provides first steps towards such a standardisation.
The library is used to
provide an extensive comparison of six different surrogate algorithms on four expensive optimisation problems from different real-life applications.
This has led to new insights regarding the relative importance of exploration, the evaluation time of the objective, and the used model. \editB{We also provide rules of thumb for which surrogate algorithm to use in which situation.}
%
A further contribution is that we make the algorithms and benchmark problem instances publicly available, contributing to more uniform analysis of surrogate algorithms. Most importantly, we include the performance of the six algorithms on all evaluated problem instances. This results in a unique new dataset that lowers the bar for researching new methods as the number of expensive evaluations required for comparison is significantly reduced.
%
%
\end{abstract}



\end{frontmatter}

\section{Introduction}

Unlike other black-box optimisation algorithms, surrogate-based optimisation algorithms such as Bayesian optimisation~\cite{Shahriari2016TakingTH} are designed specifically to solve problems with expensive objective functions.
Examples are 
materials science~\cite{Liang2020BenchmarkingTP}, temperature control~\cite{ijcai2019811},
\editA{building design~\cite{Bre2020AnEM}, 
aerodynamics~\cite{Keane2020SurrogateAF},  }optics~\cite{DONEpaper}, and computer vision~\cite{bergstra2013making}.

By making use of a surrogate model that approximates the objective function, these algorithms achieve good results even with a low number of function evaluations.
However, the training and usage of the surrogate model is more computationally intensive than the use of typical black-box optimisation heuristics such as local search or population-based methods.
This complicates thorough benchmarking of surrogate algorithms.

The current way of benchmarking surrogate algorithms does not give complete insight into the strengths and weaknesses of the different algorithms, 
\editA{such as their computational efficiency or accuracy for different types of problems.}
The most important reason \editA{for this} is the lack of a standard benchmark set of problems that come from real-life applications and that also have expensive objective functions.
\editB{This makes it difficult to determine beforehand which surrogate algorithm to use on a real-life application.}

In this work, we compare several surrogate algorithms on the same set of expensive optimisation problems from real-life applications, resulting in a public benchmark library that can be easily extended both with new surrogate algorithms, as well as with new problems.
Our other contributions are:
\begin{itemize}
    \item the creation of a meta-algorithmic dataset of surrogate algorithm performance on real-life expensive problems,
    \item insight into the strengths and weaknesses of existing surrogate algorithms \editA{depending on problem properties}, and verifying existing knowledge from literature,
    \item investigating how algorithm  performance  depends  on  the available  computational  resources  and  the  cost  of  the expensive objective,
    \item separating the effects of the choice of surrogate model and the acquisition step of the different algorithms,
    \item \editB{easy to interpret rules of thumb for when to use which surrogate algorithm}.
\end{itemize}
We furthermore show that continuous models can be used on discrete problems and vice versa.
\editA{This is an important result as in practice, the types of
variables in a problem (e.g. continuous, integer, categorical, etc.) are often used to determine which
surrogate model to choose or discard.}
The main insights that we obtained are that the accuracy of a surrogate model and the choice of using a continuous or discrete model, are less important than the evaluation time of the objective and the way the surrogate algorithm explores the search space.

\section{Background and related work} \label{sec:background} 

This section starts by giving a short explanation of surrogate-based optimisation algorithms, or surrogate algorithms for short.
We then describe some of the shortcomings in the way surrogate algorithms are currently benchmarked: the lack of standardised benchmarks and the lack of insight in computational efficiency.
Finally, we give an overview of related benchmark libraries and show how our library fills an important gap.

\subsection{Surrogate-based optimisation algorithms}

The goal of surrogate-based optimisation\editA{~\cite{Bhosekar2018AdvancesIS,hutter2011sequential,Alizadeh2020ManagingCC}} is to minimise an \emph{expensive} black-box objective function
\begin{align}
    \min_{x \in X} f(x),
\end{align}
where $X\subseteq \mathbb R^d$ is the $d$-dimensional search space with $d$ the number of decision variables.
The objective can be expensive for various reasons, but
in this work we assume $f$ is expensive in terms of computational resources, as it involves running a simulator or algorithm.
Optimising $f$ using standard black-box optimisation algorithms such as local search methods or population-based techniques may require too many evaluations of the expensive objective.
\editB{We also consider that the problem is \emph{stochastic}, meaning we only have access to noisy measurements $y_i=f(x_i)+\epsilon_i$, where the random noise variable $\epsilon_i$ is the result of randomness in the underlying simulator or algorithm.
}

Surrogate algorithms reduce the number of required objective evaluations by iterating over three steps at every iteration~$i$:
\begin{enumerate}
    \item (\emph{Evaluation}) Evaluate $y_i=f(x_i)\editB{+\epsilon_i}$ for a candidate solution $x_i$.
    \item (\emph{Training}) Update the surrogate model $g: X \rightarrow \mathbb R$ by fitting the new data point $(x_i,y_i)$.
    \item (\emph{Acquisition}) Use $g$ to determine a new candidate solution $x_{i+1}$.
\end{enumerate}

Usually, in the first $R$ iterations, $x_i$ is chosen randomly and therefore the acquisition step is skipped for these iterations.
The training step consists of machine learning techniques such as Gaussian processes or random forests, where the goal is to approximate 
the objective
$f$ with
a \emph{surrogate model} 
$g$. For the acquisition step,  an \emph{acquisition function} $\alpha$ is used that indicates which region of the search space is the most promising by trading off exploration and exploitation:
\begin{align}\label{eq:acq}
    x_{i+1} = \argmax_{x\in X} \alpha(g(x)).
\end{align}
Example acquisition functions are Expected Improvement, Upper Confidence Bound, or Thompson sampling~\cite{Shahriari2016TakingTH}.

By far the most common surrogate algorithm is Bayesian optimisation~\cite{movckus1975bayesian,Shahriari2016TakingTH}, which typically uses a Gaussian process surrogate model.
Other common surrogate models are random forests, as used in the SMAC algorithm~\cite{hutter2011sequential}, and Parzen estimators, as used in HyperOpt~\cite{bergstra2013making}.
Our own earlier work contains random Fourier features as surrogate models in the DONE algorithm~\cite{DONEpaper} and piece-wise linear surrogate models in the IDONE and MVRSM algorithms~\cite{bliek2019black,bliek2020black}.
An overview of different methods and their surrogate models is given in Table~\ref{tab:approaches}.
Details about which methods are included in the comparison are given in Section~\ref{sec:approaches}.

\begin{table*}[tb]
    \caption{Surrogate-based approaches in this benchmark environment, and whether they support continuous (cont.), integer (int.), categorical (cat.) and conditional (cond.) variables.} 
    \label{tab:approaches}
    \centering
    \begin{tabular}{llcccc}
    \toprule
    Name & Surrogate model & Cont. & Int. & Cat.  & Cond. \\
    \midrule
    SMAC\cite{hutter2010sequential-extended,hutter2011sequential} & Random forest & \checkmark  & \checkmark  & \checkmark  & \checkmark \\
    HyperOpt\cite{bergstra2013making} & Parzen estimator & \checkmark  & \checkmark  & \checkmark  & \checkmark  \\
    Bayesian Opt.\cite{Shahriari2016TakingTH,bayesianoptimization} & Gaussian process (GP) & \checkmark  & & & \\
    CoCaBO~\cite{ru2019bayesian} & GP+multi-armed bandit & \checkmark  & & \checkmark  & \\
    DONE\cite{DONEpaper} & Random Fourier  & \checkmark  & & & \\
    IDONE\cite{bliek2019black} & Piece-wise linear & & \checkmark  & & \\
    MVRSM\cite{bliek2020black} &  Piece-wise linear & \checkmark  & \checkmark  & & \\
    \bottomrule
    \end{tabular}
\end{table*}

\subsection{Shortcoming 1: lack of standardised real-life benchmarks}

Surrogate models appear to be useful to solve problems with expensive objective functions, and 
a questionnaire on real-life optimisation problems
confirms that this type of objective function often appears in practice~\cite{vanderblom2020identifying}.
%
%
Since most surrogate algorithms are developed with the goal of being applicable to many different problems, 
these algorithms should be tested on multiple benchmark functions.
Preferably, these benchmarks are \emph{standardised}, meaning that they are publicly available, easy to test on, and used by a variety of researchers.
For synthetic benchmarks, standardised benchmark libraries such as COCO~\cite{coco} have been around for several years now, 
and these types of benchmark functions are often used for the testing of surrogate algorithms as well.
However, benchmarks from real-life applications are much harder to find~\cite{Palar2019}.

Simply taking the benchmarking results on synthetic functions and applying them to expensive real-life applications, or adding a delay to the synthetic function, is not enough~\cite{Palar2019,daniels2018suite,Volz2019OnBS,Tangherloni2019BiochemicalPE,dieterich2012empirical}.
An example is the ESP benchmark discussed later in this paper.
For this benchmark we have noticed that changing only one of the variables at a time leads to no change in the objective value at all, meaning that there are more `plateaus' than in typical synthetic functions used in black-box optimisation. 
In general, expensive objectives are often expensive because they are the result of some kind of complex simulation or algorithm, and the resulting fitness landscape is therefore much harder to analyse/model than that of a synthetic function 
which can
simply be described with a mathematical function.

\subsection{Shortcoming 2: lack of insight in computational efficiency}

In many works on surrogate algorithms, computation times of the algorithms are not taken into consideration, and are often not even reported.
This is because of the underlying assumption that the expensive objective is the bottleneck.
However, completely disregarding the computation time of the surrogate algorithm leads to 
the development of algorithms that are too time-consuming to be used in practice.
In some cases, the algorithms are even slower than the objective function of the real-life application, shifting the bottleneck from the expensive objective to the algorithm.
\editB{This can be seen for example in the hyperparameter tuning problem in this work, where the hyperparameters can be evaluated faster than the slowest surrogate algorithm can suggest new values for the hyperparameters.}
Computation times should be reported, preferably for problems of different dimensions so that the scalability of the algorithms can be investigated
.
This also helps answering the open question posed in~\cite{Volz2019OnBS}: \emph{``One central question to answer is at what point
an optimisation problem is expensive ``enough'' to warrant the application
of surrogate-assisted methods.''}
Since many surrogate algorithms have a computational complexity that increases with every new function evaluation~\cite{bliek2020black}, even more preferable is to report the computation time used by the surrogate algorithm \emph{at every iteration} to gain more insight into the time it takes to run surrogate algorithms for different numbers of iterations.

Besides the computation time used by the algorithms, different real-life applications have different \emph{budgets} available that put a limit on the number of function evaluations or total computation time.
Taking this computational budget into account is a key issue when tackling real-world problems using surrogate models~\cite{Palar2019}.
Yet for most surrogate algorithms, it is not clear how they would perform for different computational budgets.

\subsection{Related benchmark environments}

From the way surrogate algorithms are currently benchmarked and the shortcomings that come with it, we conclude that we do not sufficiently understand the performance regarding both quality and run-time on realistic expensive black-box optimisation problems.
A \emph{benchmark library} can help in gaining more insight as algorithms are compared on the same set of test functions.
In the context of black-box optimisation, such a library consists of multiple objective functions and their details (such as 
the number of continuous or integer variables, evaluation time, etc.) and possibly of baseline algorithms that can be applied to the problems.
For non-expensive problems, many such libraries exist~\cite{coco,Wagner2005HeuristicLabAG,Humeau2013ParadisEOMOFF}, particularly with synthetic functions.
Some of these libraries also contain real-life functions that are not expensive~\cite{ochoa2012hyflex,Caraffini2020TheSP,IOHprofiler}.
See Table~\ref{tab:benchmarkenvs} for an overview of related benchmark environments.

The real-life problems to which surrogate algorithms are usually applied can roughly be divided into computer science problems and engineering problems, or digital and physical problems. Examples of the former are algorithm configuration problems~\cite{hutter2011sequential}, while the latter deal with (simulators of) a physical problem such as aerodynamic optimisation~\cite{Liu2012COMPARISONOI}.
Even though surrogate models are used in both problem domains, these two communities often stay separate: 
most benchmark libraries that contain \emph{expensive} real-life optimisation problems only deal with one of the two types, for example in automated machine learning~\cite{amlb2019,Ying2019NASBench101TR,Dong2020NASBench201ET,siems2020nasbench301}
or computational fluid dynamics~\cite{daniels2018suite}.
The problem with focusing on only one of the two domains is that domain-specific techniques such as early stopping of machine learning algorithms~\cite{BOHB} or adding gradient information from differential equations~\cite{Han2013ImprovingVS,SMT2019} are exploited when designing new surrogate algorithms, making it difficult to transfer the domain-independent scientific progress in surrogate algorithms from one domain to the other.
\editB{Benchmarking surrogate algorithms in multiple problem domains would be beneficial for all these domains.}

Though all benchmark libraries contain benchmark problems, not all of them contain
\emph{solutions} in the form of surrogate algorithms, and some of them do not even contain any type of solution at all.
One library that does contain many surrogate algorithms is SUMO~\cite{sumo}, a commercial toolbox 
with a wide variety of applications both in computer science and engineering.
Unfortunately, this Matlab tool is over $10$ years old, and only a restricted version is available for researchers, making it less suitable for benchmarking.
It only supports low-dimensional continuous problems, and newer surrogate algorithms that were developed in the last decade are not implemented.

What is currently missing is a modern benchmark library that is aimed at real-life expensive benchmark functions not just from computer science but also from engineering, and that also contains baseline surrogate algorithms that can easily be applied to these benchmarks such as SMAC, HyperOpt, and Bayesian optimisation with Gaussian processses.








\section{Proposed Benchmark Library: \bname}\label{sec:proposed}

In this section we introduce \bname: an EXPensive Optimisation benchmark library%
\footnote{\editB{Our code is available publicly at \url{\coderepositoryurl}}}%
.
We propose a benchmark suite focusing on single-objective, expensive, real-world problems, consisting of many integer, categorical, and continuous variables or mixtures thereof.
The problems come from different engineering and computer science applications, and we include seven baseline surrogate algorithms to solve them.
See Table~\ref{tab:benchmarkenvs} for details on how \bname~ compares to related benchmark environments.


\begin{table}[tb]
    \centering
    \caption{Related benchmark environments
    }
    \label{tab:benchmarkenvs}
    \begin{tabular}{p{0.21\columnwidth}cccc}
    \toprule
    Name & \begin{minipage}{0.14\columnwidth}{Contains expensive problems}\end{minipage} & \begin{minipage}{0.12\columnwidth}{Contains engineering problems}\end{minipage} & \begin{minipage}{0.12\columnwidth}{Contains computer science problems}\end{minipage} & \begin{minipage}{0.15\columnwidth}{Implemented surrogate algorithms}\end{minipage}\\
    \midrule
    HyFlex~\cite{ochoa2012hyflex}
    & & & \checkmark & 0\\
    SOS~\cite{Caraffini2020TheSP}
    & & \checkmark & & 0\\
    IOHprofiler~\cite{IOHprofiler}
    & & \checkmark & \checkmark & $0$ \\
    GBEA~\cite{Volz2019SingleAM}
    & \checkmark & & \checkmark & 0\\
    CFD~\cite{daniels2018suite}
    & \checkmark & \checkmark & & 0\\
    \begin{minipage}{1.3\columnwidth}
    \mbox{NAS-Bench}~\cite{Ying2019NASBench101TR,Dong2020NASBench201ET,siems2020nasbench301}
    \end{minipage}
    & \checkmark & & \checkmark & 0\\
    \mbox{DAC-Bench}~\cite{eimer-ijcai21}
    & \checkmark & & \checkmark & 0\\
    RBFopt~\cite{Costa2018RBFOptAO}
    &  & \checkmark & & $1$\\
    CompModels~\cite{Pourmohamad2020CompModelsAS}
    & \checkmark & \checkmark & & 1\\
    HPObench~\cite{hpolib}
    & \checkmark 
    & & \checkmark & $2$\\
    AClib2~\cite{aclib2}
    & \checkmark & & \checkmark & 2\\
    Nevergrad~\cite{nevergrad}
    & \checkmark & \checkmark & \checkmark & $2$\\
    BayesMark~\cite{bayesmark}
    & \checkmark & & \checkmark & $3$\\
    MATSuMoTo~\cite{matsumoto}
    &  &  &  & $4$\\
    AMLB~\cite{amlb2019}& \checkmark & & \checkmark & $4$\\
    PySOT~\cite{eriksson2019pysot}
    & & & & 5\\
    \textbf{\bname} & \checkmark & \checkmark & \checkmark & $\mathbf{7}$\\
    SUMO~\cite{sumo}
    & \checkmark & \checkmark & \checkmark & $9$\\
    SMT~\cite{SMT2019}
    &  & \checkmark & & 14\\
    \bottomrule
    \end{tabular}
    
\end{table}

The simple framework of this benchmark library makes it possible for researchers in surrogate models to compare their algorithms on a standardised set of real-life problems, while researchers with expensive optimisation problems can easily try a standard set of surrogate algorithms on their problems.
This way, our benchmark library advances the field of surrogate-based optimisation.

It should be noted that synthetic benchmark functions are still useful, as they are less time-consuming and have known properties.
We therefore still include synthetic benchmarks in our library, though we do not discuss them in this work.
We encourage researchers in surrogate models to use synthetic benchmarks when designing and investigating their algorithm, and then use the real-life benchmarks presented in this work as a stress test to see how their algorithms hold up against more complex and time-consuming problems.

In the remainder of this section, we describe the problems and the approaches to solve these problems that we have added to \bname.


\subsection{Included Expensive Benchmark Problems}

The problems that were included in \bname\ were selected in such a way that they contain a variety of applications, dimensions, and search spaces.
To encourage the development of surrogate algorithms for applications other than computer science, we included several engineering problems, one of which was first introduced in the CFD benchmark library~\cite{daniels2018suite}.
\editB{Many surrogate algorithms claim to work on a wide variety of expensive optimisation problems. However, benchmarking is often limited to synthetic problems, or real-life problems from one domain, casting doubt on the validity of these claims. To verify the general usefulness of surrogate algorithms, it is important to test them on problems from different domains.}

The problem dimensions \editB{in this work} were chosen to be difficult for standard surrogate algorithms: Bayesian optimisation \editB{with Gaussian processes} is typically applied to problems with less than $10$ variables.
Two of our problems have $10$ variables, though it is possible to scale them up, while the other problems contain tens or even over $100$ variables.
This is in line with our view of designing surrogate algorithms using easy, synthetic functions, and then testing them on more complicated real-life applications.
Since \emph{discrete} expensive problems are also an active research area, we included one discrete problem and even a problem with a mix of discrete and continuous variables.

The problems were carefully selected to have expensive objectives that take longer to evaluate than synthetic functions, but not so long that benchmarking becomes impossible. 
On our hardware (see Section~\ref{sec:hardware}), the time it takes to evaluate the objective function varies between $2$ and $60$ seconds depending on the problem.
\editB{However, all included benchmarks capture the properties of even more expensive problems, or can be made more expensive by changing the corresponding data or problem parameters. We also provide a method to simulate situations where evaluating the objective requires significantly more time.}

\editA{By the nature of \editB{the chosen} problems, there are no preconstructed and unrealistic relations between variables. Hence, the additional complexity stemming from real-world objectives makes the benchmarking more interesting as well, such as the aforementioned greater magnitude of `plateaus' in the ESP problem compared to typical synthetic functions.  Additionally, since these objectives are more difficult to analyse, 
it is harder to come up with a solver that exploits any knowledge of the problem.}

We now give a short description of the four real-life expensive optimisation problems that are present in \bname. \editB{This information is summarised in Table~\ref{tab:problems}.
}



\begin{table}[tb]
    \centering
    \caption{\editB{Included expensive optimisation problems, and the approximate evaluation time of the objective (eval.), dimension or total number of variables (dim.), corresponding number of continuous (cont.), integer (int.), and categorical (cat.) variables, and whether some variables are conditional (cond.) variables.}
    }
    \label{tab:problems}
    \begin{tabular}{lcccccc}
    \toprule
    Problem & Eval. & Dim. & Cont. & Int. & Cat. & Cond. \\
    \midrule
    Windwake & $15s$ & $10$ & $10$ & - & - & -\\
    Pitzdaily & $2$-$60s$ & $10$ & $10$ & - & - & -\\
    ESP & $28s$ & $49$ & - & - & $49$ & -\\
    HPO & $1$-$8s$ & $135$ & $11$ & $7$ & $117$ & yes\\
    \bottomrule
    \end{tabular}
\end{table}

\subsubsection{Wind Farm Layout Optimisation (Windwake)}
This benchmark utilises a wake simulator called FLORIS~\cite{floris2020} to determine the amount of power a given wind farm layout produces.
To make the layout more robust to different wind conditions, we decided to use as output 
the power averaged over multiple scenarios, where each scenario uses randomly generated wind rose data, generated with the same distribution.
%
A solution is represented by a sequence of pairs of coordinates for each wind turbine, which can take on continuous values.
The output is $-1$ times the power averaged over multiple scenarios, which takes about $15$ seconds to compute on our hardware for $5$ scenarios and $5$ wind turbines.
It should be noted that this particular problem has constraints besides upper and lower bounds for the position of each wind turbine: turbines are not allowed to be located within a factor of two of each others’ radius. 
\editA{This is not just for modelling a realistic situation: the simulator fails to 
provide accurate results if this overlap is present.
With the relatively low packing density present in this problem, this gives a realistic and interesting fitness landscape.
}
As the goal of this work is not to compare different ways to handle constraints, we use the naive approach of incorporating the constraint in the objective.
The objective simply returns $0$ when constraints are violated.
\editB{While more complex wake simulators exist, the problem also becomes more expensive when the number of wind turbines and the number of scenarios are increased.}

\subsubsection{Pipe Shape Optimisation (Pitzdaily)}

One of the engineering benchmark problems proposed in the CFD library~\cite{daniels2018suite}, called PitzDaily, is pipe shape optimisation.
This benchmark uses a computational fluid dynamics simulator to calculate the pressure loss for a given pipe shape.
The pipe shape can be specified using $5$ control points, giving $10$ continuous variables in total.
The time to compute the pressure loss varies from $2$ to $60$ seconds on our hardware.
Although the search space is continuous, there are constraints to this problem: violating these constraints returns an objective value of $2$, which is higher than the objective value of feasible solutions.
\editB{This problem becomes more expensive with an increase in the number of control points.}

\subsubsection{Electrostatic Precipitator (ESP)}

This engineering benchmark contains only discrete variables.
The ESP is used in industrial gas filters to filter pollution.
The spread of the gas is controlled by metal plates referred to as baffles. Each of these baffles can be solid, porous, angled, or even missing entirely. This categorical choice of configuration for each baffle constitutes the search space for this problem.
There are $49$ baffle slots in total, that each have $8$ categorical options.
The output is calculated using a computational fluid dynamics simulator~\cite{rehbach2018comparison}, which takes about $28$ seconds to return the output value on our hardware.
\editB{The underlying simulator becomes more expensive when using a more fine-grained simulation mesh~\cite{rehbach2018comparison}.}

\subsubsection{Hyperparameter Optimisation and Preprocessing for XGBoost (HPO)}

This automated machine learning benchmark is a hyperparameter optimisation problem. The approach, namely an XGBoost~\cite{chen2016xgboost} classifier, has already been selected.
\editB{It is one of the most common machine learning models for tabular data.
The model} contains a significant number of configuration parameters of various types, including parameters on the pre-processing step. Variables are not only continuous, integer or categorical, but also conditional: some of them remain unused depending on the value of other variables.
In total, there are $135$ variables, most of which are categorical.
The configuration is evaluated by 5-fold cross-validation on the Steel Plates Faults dataset\footnote{\url{http://archive.ics.uci.edu/ml/datasets/Steel+Plates+Faults}}, and the output of the objective uses this value multiplied with $-1$.
Since there can be a trade-off between accuracy and computation time for different configurations, we set a time limit of $8$ seconds, as this was roughly equal to twice the time it takes to use a default configuration on our hardware.
Configurations for which the time limit is violated, return an objective value of $0$.
\editB{This problem becomes more expensive when using a larger dataset that requires a longer training time.
}

\subsection{Approaches}\label{sec:approaches}


In this section we show the approaches that are considered in the benchmark library.
We limit ourselves to popular single-objective surrogate algorithms \editA{due to the limited evaluation budget that usually accompanies an expensive objective function. Furthermore, the approaches are} easily implemented and open-source, and do not focus on extensions of the expensive optimisation problem such as a batch setting, multi-fidelity or multi-objective setting, highly constrained problems, etc.
These include a Bayesian optimisation algorithm~\cite{Shahriari2016TakingTH,bayesianoptimization}, which uses Gaussian processes with a Mat\'ern 5/2 kernel \editA{and upper confidence bound acquisition function ($\beta=2.576$)}, 
SMAC~\cite{hutter2011sequential}, and HyperOpt\cite{bergstra2013making}.
We also include our own earlier work~\cite{DONEpaper,bliek2019black,bliek2020black}, with the DONE, IDONE and MVRSM algorithms.
A recent variant of Bayesian optimisation, namely CoCaBO~\cite{ru2019bayesian}, is also included in the benchmark library, but not presented in this work due to the required computation time.
The baseline with which all algorithms are compared is random search~\cite{Bergstra2012RandomSF}, for which we use HyperOpt's implementation.
We also include several local and global search algorithms in our library (Nelder-Mead, Powell's method and basin-hopping among others), but these failed to outperform random search on all of our benchmark problems, and are therefore not presented in this work.

Not all of these algorithms can deal with all types of variables, although often naive implementations are possible: discretisation to let discrete surrogates deal with continuous variables, rounding to let continuous surrogates deal with discrete variables, and/or ignoring the conditional aspect of a variable entirely.
Table~\ref{tab:approaches} shows the types of variables that are directly supported by the surrogate models used in each algorithm.

\section{Results}\label{sec:results}

The different surrogate algorithms are objectively compared on all four different real-life expensive benchmark problems of \bname.
The goals of the experiments are 
to 1) gain insight into the strengths and weaknesses of existing surrogate algorithms and verify existing knowledge from literature, 2) investigate how algorithm performance depends on the available computational resources and the cost of the expensive objective, and 3) separate the effects of the choice of surrogate model and the acquisition step of the different algorithms.

The results of comparing the different surrogate algorithms on the problems of \bname\ provide a new dataset that we use for these three goals, and that we make available publicly.%
\footnote{The dataset can be found online at \url{https://doi.org/10.4121/14247179}}

This dataset includes the points in the search space chosen for evaluation by each algorithm, the resulting value of the expensive objective, the computation time used to evaluate the objective, and the computation time used by the algorithm to suggest the candidate point.
The latter includes both the 
training and acquisition steps of the algorithms,
as it was not easy to separate these two for all algorithms.
Although we perform some initial analysis on this meta-algorithmic dataset, it can also be used by future researchers in, for example, instance space analysis~\cite{SMITHMILES201412}
or building new surrogate benchmarks from this tabular data~\cite{siems2020nasbench301}.

We start this section by
giving the experimental details, followed by the results on the four benchmark problems.
We then investigate the influence of the computational budget and cost of the expensive objective, followed by a separate investigation of the choice of surrogate model.
\editB{The section ends with a summary of the obtained insights.}


\subsection{Experiment details\label{sec:hardware}}
\subsubsection{Hardware}
We use the same hardware when running the different surrogate algorithms on the different benchmark problems.
All these experiments are performed in Python, on a {Intel(R) Xeon(R) Gold 6148 CPU @ 2.40GHz} with 32 GB of RAM. Each approach and evaluation was performed using only a single CPU core.

\subsubsection{Hyperparameter settings}

All methods use their default hyperparameters with the exception of SMAC, which we set to deterministic mode to avoid repeating the exact same function evaluations, which drastically decreased performance in our experience.
For the MVRSM method, we set the number of basis functions in purely continuous problems to $1000$.
We have not adapted IDONE for continuous or mixed problems.

\subsubsection{Normalisation}
To make comparison between benchmarks easier, we normalise the best objective value found by each algorithm at each iteration in the figures shown in this section.
This is done as follows: using the best objective value found by random search as a baseline, let $r_0$ be the average of this baseline after $1$ iteration, and let $r_1$ be the average of this baseline after the number of random initial guesses $R$ that each algorithm used\footnote{\editA{Note: 
we used the same uniform distribution for the random initial guesses of all methods, and new samples were drawn at every run to remove dependence on the initialisation. The samples themselves are different across algorithms.
}}%
.
Then all objective values $f$ are normalised as
\begin{align}
    f_{norm} = (f-r_0)/(r_1-r_0),
\end{align}
meaning that $r_0$ corresponds to a normalised objective of $0$ and $r_1$ corresponds to a normalised objective of $1$, and a higher normalised objective is better.
Note that this is only used in Figure~\ref{fig:4subs}.
This normalisation is possible since all surrogate algorithms start with the same number of random evaluations $R$, which we omit from the figures.
Other visualisation tools that are popular in black-box optimisation, such as ECDF curves, are less suitable for our results since the optimum is unknown for our benchmarks and we look at only one benchmark at a time.
\editA{Another metric, namely the area under the curve, is shown in \ref{sec:aoc}.}

\subsubsection{Software environment}


\bname~is available as a public github repository\footnote{\url{\coderepositoryurl}} and is implemented in the Python programming language.
To stimulate future users to add their own problems and approaches to this library, we have taken care to make this as easy as possible and provide documentation to achieve this.
We also provide an interface that can easily run one or multiple approaches on a problem in the benchmark suite using the command line interface in \texttt{run\_experiment.py}.
An example is the following code:
\begin{spverbatim}
python run_experiment.py --repetitions=7 --out-path=./results/esp 
--max-eval=1000 --rand-evals-all=24 esp 
randomsearch hyperopt bayesianoptimization
\end{spverbatim}
This runs random search, HyperOpt and Bayesian optimisation on the ESP problem for $1000$ iterations, of which the first $24$ iterations are random, repeated seven times, and outputs the results in a certain folder.

\subsection{Benchmark results}
We now share the results of applying all algorithms in \bname \ to the different benchmark problems. 
The IDONE algorithm is only applied to the ESP problem since it does not support continuous variables.
To investigate statistical significance of the results, we also report p-values of a pair-wise Student's T-test at the last iteration on unnormalised data.

\begin{figure*}[p]
    \centering
    \subfloat[]{
         \centering
         \includegraphics[width=0.88\textwidth]{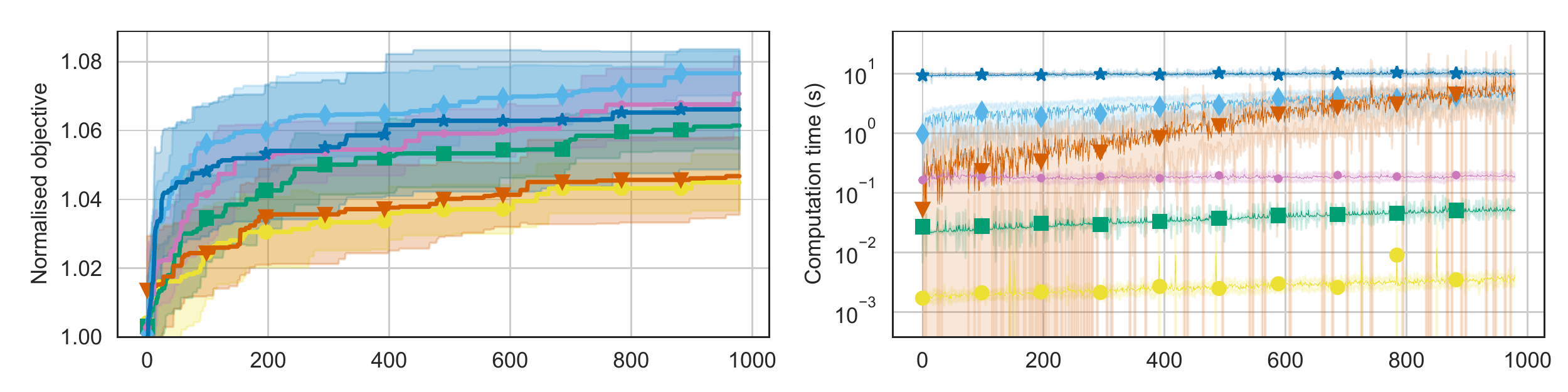}
         \label{fig:WFLO}
         }\\
    \subfloat[]{
        \centering
        \includegraphics[width=0.88\textwidth]{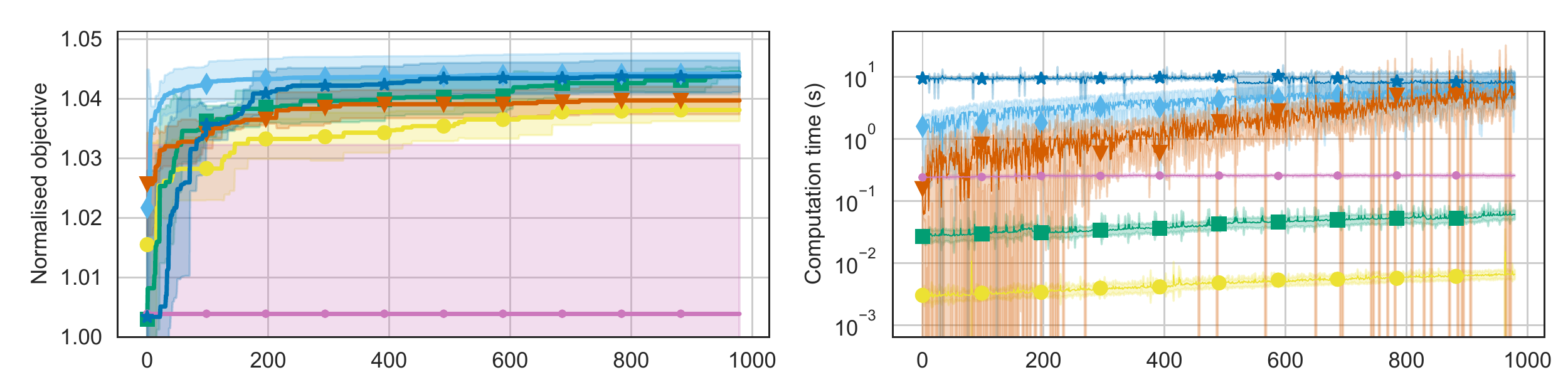}
        \label{fig:Pitz}
    }\\
    \subfloat[]{
        \centering
        \includegraphics[width=0.88\textwidth]{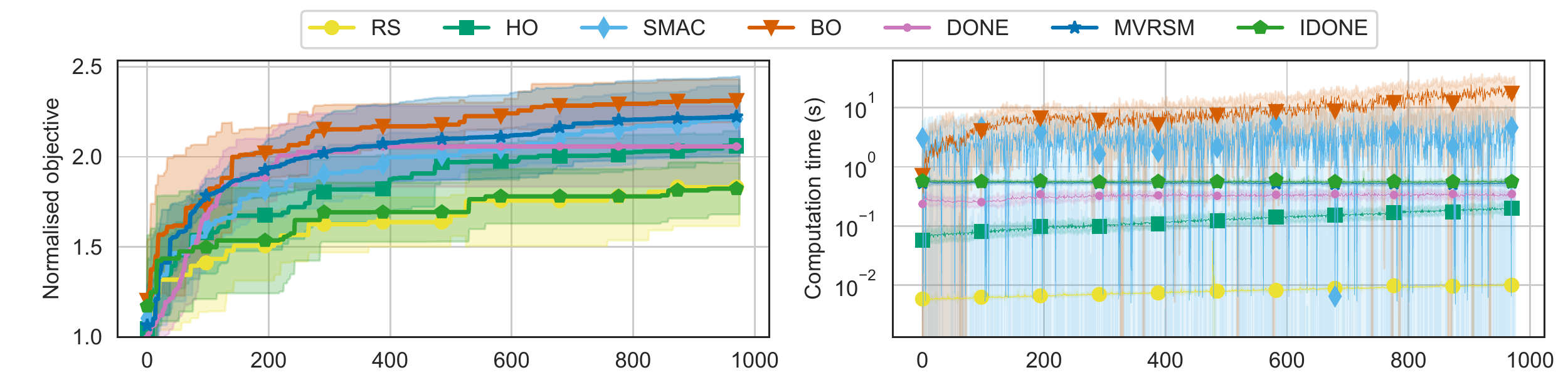}
        \label{fig:ESP}
        }\\
    \subfloat[]{
        \centering
        \includegraphics[width=0.88\textwidth]{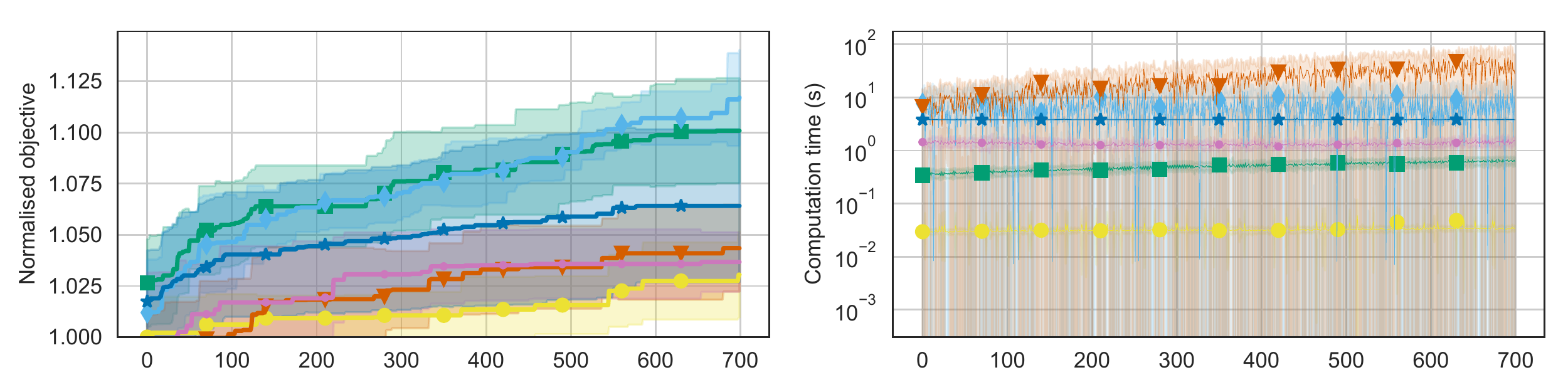}
        \label{fig:HPO}
        }
    \caption{Results on the different benchmark problems, averaged over $T$ runs, after starting with $R$ random samples. \editB{$T$ is varied due to the different degrees of expensiveness of the problems.} The shaded area indicates one standard deviation, the horizontal axis indicates the iteration of the algorithm, and all figures use the legend shown in the middle.
    The computation time on the right does not contain the time it takes to evaluate the objective.
    The benchmark problems are: (a) wind farm layout optimisation, $10$ continuous variables, $T=10$, $R=20$; (b) Pitzdaily, $10$ continuous variables, $T=5$, $R=20$;
    (c)  electrostatic precipitator, $49$ discrete variables, $T=7$, $R=24$; 
    (d) simultaneous hyperparameter tuning and preprocessing for XGBoost, $117$ categorical, $7$ integer, $11$ continuous variables, $T=10$, $R=300$.
    }
    \label{fig:4subs}
\end{figure*}

\subsubsection{Windwake}
For the wind farm layout optimisation problem, 
Figure~\ref{fig:WFLO} shows the normalised best objective value found at each iteration by the different algorithms, as well as the computation time used by the algorithms at every iteration.
All algorithms started with $R=20$ random samples not shown in the figure.
None of the algorithms use more computation time than the expensive objective itself, which took about $15$ seconds per function evaluation.
While random search is the fastest method, it fails to provide good results, as is expected for a method that does not use any model or heuristic to guide the search.
Interestingly, Bayesian optimisation (BO) does not outperform random search on this problem ($p>0.6$
) and is outperformed by all other methods ($p<0.01$), even though it is designed for problems with continuous variables.
In contrast, MVRSM and SMAC both have quite a good performance on this problem while they are designed for problems with mixed variables, though they both do take up more computational resources.
DONE, another algorithm designed for continuous problems, performs similar to MVRSM and SMAC ($p>0.1$).



\subsubsection{Pitzdaily}

Figure~\ref{fig:Pitz} shows the results of the Pitzdaily pipe shape optimisation problem with $R=20$.
It can be seen that DONE fails to provide meaningful results.
Upon inspection of the proposed candidate solutions, it turns out that the algorithm gets stuck on parts of the search space that violate the constraints.
This happens even despite finding feasible solutions early on and despite the penalty for violating the constraints.
SMAC, HyperOpt (HO) and MVRSM are the best performing methods on average, outperforming the other three methods ($p<0.05$) but not each other ($p>0.6$
).


\subsubsection{ESP}

In this discrete problem, algorithms that only deal with continuous variables resort to rounding when calling the expensive objective.
This is considered suboptimal in literature, however earlier work shows that this is not necessarily the case for the ESP problem~\cite{BNAIC2020paper}.
Indeed, we see in Figure~\ref{fig:ESP} that Bayesian optimisation is the best performing method on this problem, outperforming all methods ($p<0.03$) except MVRSM and SMAC ($p>0.2$
).
This counters the general belief that Bayesian optimisation with Gaussian processes is only adequate on low-dimensional problems with only continuous variables.
Another observation is that MVRSM performs much better than IDONE ($p<0.01$
), which fails to significantly outperform random search ($p>0.9$
) even though IDONE is designed for discrete problems.
The surrogate algorithms also use less computation time than the expensive objective which took about $28$ seconds per iteration to evaluate.




\subsubsection{XGBoost Hyperparameter Optimisation}

Like in the previous benchmark, the algorithms that only deal with continuous variables use rounding for the discrete part of the search space in this problem.
For dealing with conditional variables with algorithms that do not support them we use a naive approach: changing such a variable simply has no effect on the objective function when it disappears from the search space, resulting in a larger search space than necessary.
Figure~\ref{fig:HPO} shows the results for this benchmark.
This time, results are less surprising as SMAC and HyperOpt, two algorithms designed for hyperparameter optimisation with conditional variables, give the best performance.
Though they perform similar to each other ($p>0.1$
), they outperform all other methods ($p<0.03$).
MVRSM is designed for mixed-variable search spaces like in this problem, but not for conditional variables, and outperforms random search ($p<0.03$
) but not BO and DONE ($p>0.05$
).
BO and DONE both fail to outperform random search ($p>0.1$
).
If we also consider computation time, HyperOpt appears to be a better choice than SMAC, being faster by more than an order of magnitude, \editB{while BO is even slower than SMAC}.

\subsection{Varying time budget and function evaluation time}


In this experiment we investigate how the algorithms perform with various time budgets and different objective evaluation times. More specifically, instead of restricting the number of evaluations as done up until now, the algorithms are stopped if their runtime exceeds a fixed time budget. \editB{This} runtime includes both the total function evaluation time as well as the computation time required for the training and acquisition steps of the algorithm.  
This experiment extends the results of the benchmark by putting emphasis on the computation time of the algorithm in addition to their respective sample efficiency. On top of that, it provides information that can be used to decide which algorithm is suitable given a time budget and how expensive the objective function is.  

To investigate this in practice, we use the data gathered in the experiments shown in this section by artificially changing the time budget and evaluation time of the expensive objective functions as in earlier work~\cite{BNAIC2020paper}. Because we know the computation times from each iteration in the experiments, it is possible to simulate what the total runtime would be if the function evaluation time is adjusted. Then, we report which algorithm returns the best solution when the time budget has been reached for various time budgets and evaluation times. 
The evaluation time ranges from  $0.12$ ms to $36$ hours, while the time budget ranges from $0.49$ ms to $36$ hours.
In case the time budget is not reached within the maximum number of iterations that we have observed from the other experiments, for at least one of the algorithms, no results are reported.


\begin{figure*}[tb]
\centering
    \includegraphics[width=0.92\textwidth]{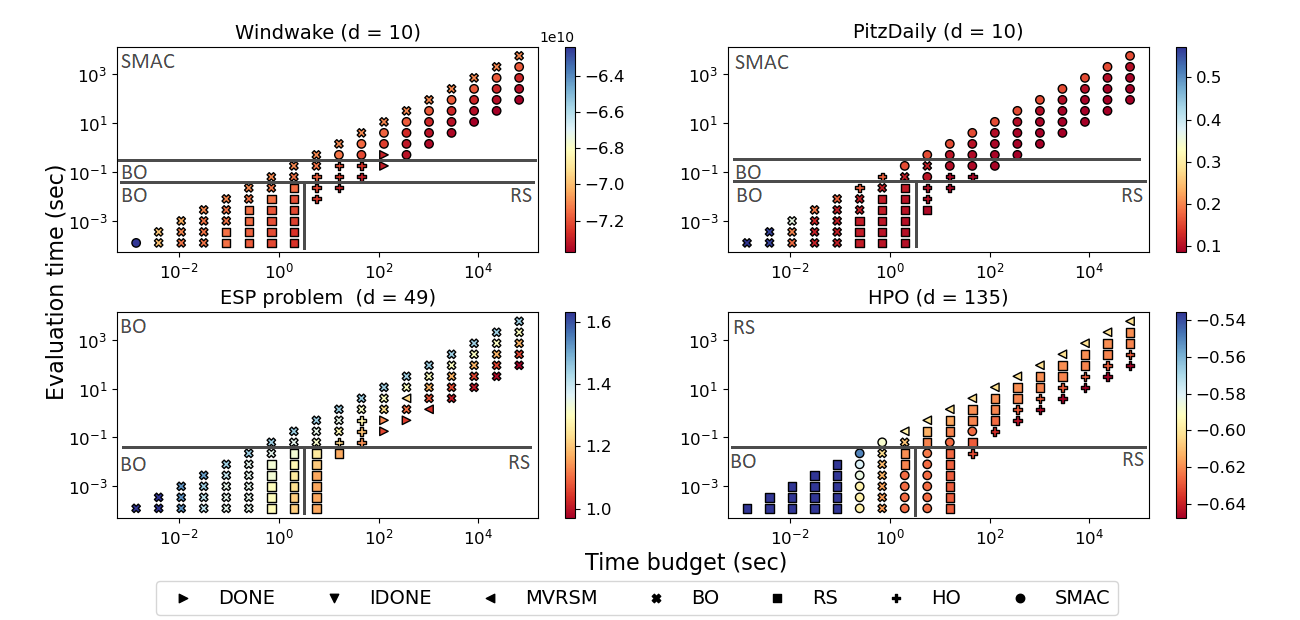}
    \caption{The best surrogate algorithm for the case that the evaluation time of the objective is artificially changed (vertical axis), and for different time budgets (horizontal axis). The different marker shapes indicate which of the surrogate algorithms achieved the best objective value, while the colour shows the corresponding objective value (not normalised, \editB{lower is better}).
    \editB{The regions divided by black lines show which algorithm would perform best according to a decision tree trained on the data.}
    Other black-box optimisation algorithms such as population-based methods are expected to dominate the empty bottom right region, where the time budget is large but the function evaluation time is small.}
    \label{fig:2d}
\end{figure*}

Figure~\ref{fig:2d} displays which algorithm returns the best solution at each problem for a variety of time budgets (x-axis) and function evaluation times (y-axis). Each algorithm has a different marker, and the colour indicates the objective value of the best found solution (without normalisation, so lower is better). As expected, we observe that the objective value decreases
when the time budget increases and the evaluation time remains fixed. However, it appears that different algorithms perform well in regions with certain time budgets and evaluation times. 

For the Windwake problem we see that 
almost all algorithms
perform the best in different settings.
BO seems to perform best when the number of iterations is low no matter the time budget, SMAC performs best for larger time budgets and evaluation times, and random search performs best for low evaluation times. HyperOpt and DONE perform well on semi-expensive objective functions in the $10-1000$ ms range. 
%
%
The observations are similar for the PitzDaily and ESP problems, except that DONE had a poor performance on the PitzDaily problem and SMAC gets outperformed by BO on the ESP problem.
%
%
Lastly, for the hyperparameter optimisation problem, it can be seen that HyperOpt is favoured over SMAC due to its computational efficiency, though SMAC performs well with cheaper objective functions.
Given a low enough time budget, random search  gives the best results, even for expensive objective functions. 


\subsubsection{\editB{Rules of thumb}}\label{sec:thumb}
\editB{
To turn these insights into easily interpretable rules of thumb, we have trained a decision tree classifier on the data points of Figure~\ref{fig:2d}, and the corresponding decision rules are indicated by regions separated by black lines. The decision tree takes as input the time budget and evaluation time, and two problem features: one feature that indicates whether the problem is a $10$-dimensional continuous problem (true for Windwake and PitzDaily, false for ESP and HPO), and a feature that indicates whether the problem contains a computational fluid dynamics (CFD) simulator (true for PitzDaily and ESP, false for ESP and HPO). These features were chosen to prevent problem-specific features: now, at least two features are needed to get a decision for one specific problem. The class label output of the decision tree is the best surrogate algorithm according to the data, which it was able to predict with a training accuracy of $0.63$ and a test accuracy of $0.71$ after a $80\%-20\%$ train-test split and repeating the training procedure $10$ times and keeping the tree with the highest test accuracy. This procedure took less than one second of computation time. The decision tree had a maximum depth of $5$ and a maximum of $6$ leaf nodes, while otherwise using default hyperparameter settings from Python's scikit-learn package.}

\editB{
The resulting decision tree led to the following rules of thumb for which surrogate algorithm to choose: 
\begin{enumerate}
    \item For cheap objective functions (at most tens of milliseconds evaluation time) with a tight time budget (around 1 second or less), BO is a good option, with random search as a close second.
    \item For cheap objective functions with a high time budget (at least seconds), random search is the best option, meaning no surrogate algorithm is required. This is the typical setting for black-box optimisation, so we expect many algorithms to outperform surrogate algorithms in this case.
    \item For expensive high-dimensional non-continuous problems that do not make use of a CFD simulator (like hyperparameter optimization), random search is a good option, with HyperOpt and MVRSM as close alternatives.
    \item For expensive high-dimensional non-continuous problems that make use of a CFD simulator (like ESP), BO is the best option.
    \item For semi-expensive (tens to hundreds of milliseconds) continuous 10-dimensional problems, BO is a good option, with SMAC and HyperOpt as close alternatives.
    \item For expensive continuous 10-dimensional problems, SMAC is the best option.
\end{enumerate}}

\editB{
Together, these rules of thumb can give practical insights which would otherwise require familiarity with the literature on surrogate algorithms.
Besides this, while most of these rules of thumb are in line with existing literature, rules 4 and 6 actually oppose existing knowledge.
This is because BO in this work makes use of a Gaussian process, which is often claimed to work well only on continuous problems with $20$ variables or less, see e.g.}~\cite{Moriconi2020HighdimensionalBO,combo}\editB{, while rule 4 shows it works well on high-dimensional non-continuous problems such as ESP. On the other hand, rule 6 shows that SMAC, which uses a random forest surrogate, works well mostly for continuous 10-dimensional problems, while random forests are well-suited for high-dimensional non-continuous search spaces.}

\editB{
Finally, several problem settings have been identified where random search shows a strong performance, such as those in rules 1, 2 and 3. It is likely that other black-box optimisation algorithms that do not make use of a surrogate, such as evolutionary algorithms, would outperform surrogate-based methods in these cases, though more research is necessary to verify this.}

\subsection{Offline learning of surrogates}\label{sec:offline}


As a final experiment we investigate the choice of surrogate model in the different surrogate algorithms.
We show how the dataset generated in this work can be used in a simple offline supervised learning framework
by training and testing different models on the data and considering the resulting errors.
We limit the scope to the Pitzdaily and ESP problems here, and generate different training sets for each (more data, as well as standard deviations, can be found in \ref{app:offline}).
Each training set consists of the first $500$ candidate solutions and objective function values gathered by one run of a specific algorithm, including the first random iterations.
We then train a variety of machine learning models on this dataset, with the goal of predicting the (unnormalised) objective function value corresponding to the candidate solution.
Using a quadratic loss function, this results in a number of machine learning models equal to the number of algorithms times the number of runs, for each type of machine learning model.
The models we used are taken from the Python scikit-learn library~\cite{scikit-learn}, and we also add XGBoost and the piece-wise linear model used by the IDONE and MVRSM algorithms, giving the following models: linear regression model (Linear), piece-wise linear model (PWL), random forest with default hyperparameters (RF), XGBoost with default hyperparameters (XGBoost), and the Gaussian process used by Bayesian optimisation (GP). 

As a test set we concatenate all the candidate points and function evaluations that were evaluated by each surrogate algorithm for every run, and keep the $1000$ points with the best objective value for each problem.
As the global optimum is unknown in these problems, this shows how the different models would perform in good regions of the search space.

Table~\ref{tab:offlineRS} shows the results of training each model on data gathered by random search, averaged over different runs.
We can immediately see that some models are prone to overfitting: the models with the smallest training errors are not necessarily the most accurate near the optimum, and may even be outperformed by a simple linear regression model there.
Furthermore, discrete models such as random forest and XGBoost with default hyperparameters have a good generalisation performance, not just on the discrete ESP problem but also on the continuous Pitzdaily problem, even though their training error is a bit higher than that of other models.

If we train models on data gathered by a surrogate algorithm that uses that model or an approximation thereof, we get the results shown in Table~\ref{tab:offlineSame}.
The models PWL and GP are exactly the same as the ones used in the corresponding surrogate algorithms (IDONE/MVRSM and BO respectively), while SMAC uses a random forest with different hyperparameters than the RF model used here, and DONE only uses an approximation of a Gaussian process.
The training error on data gathered by DONE can get very low, but this does not mean that DONE is a good surrogate algorithm, as we saw it perform poorly on the Pitzdaily problem.
A likely explanation is that the acquisition is not leading to the right data points.
More interesting are the test errors: though the GP trained on data gathered by a surrogate algorithm that uses this model (BO) receives a low test error, an XGBoost model trained on data gathered by random search can get an even lower test error; see Table~\ref{tab:offlineRS}.
The test error for XGBoost trained on data gathered by BO, not shown in these tables, is $0.997$ for the Pitzdaily problem and $0.701$ for the ESP problem. 

\begin{table}[tb]
    \centering
    \caption{Mean average error for models trained on data gathered by random search, averaged over different random search runs.}
    \label{tab:offlineRS}
    \begin{tabular}{ccccc}
    \toprule
    Benchmark & 
    \multicolumn{2}{c}{Pitzdaily} & \multicolumn{2}{c}{ESP}\\
    \cmidrule(r){1-1}\cmidrule(lr){2-3}\cmidrule(lr){4-5}
    Model & 
    Train & Test & Train & Test\\
    \midrule
    Linear &
    $0.697$ & $1.229$ &
    $2.651$ & $1.808$ \\
    PWL & 
    $0.006$ & $1.323$ & 
    $\mathbf{5\cdot 10^{-9}}$ & $8.869$ \\
    RF &
    $0.151$ & $\mathbf{0.758}$ & 
    $0.574$ & $0.972$ \\
    XGBoost &
    $0.279$ & $0.849$ &
    $0.153$ & $\mathbf{0.711}$ \\
    GP & 
    $\mathbf{8\cdot 10^{-7}}$ & $1.147$ &
    $1\cdot 10^{-6}$ & $1.297$ \\
    \bottomrule
    \end{tabular}
\end{table}

\begin{table}[tb]
    \centering
    \caption{Mean average error for models trained on data gathered by a surrogate algorithm that uses that model, averaged over different runs. }
    \label{tab:offlineSame}
    \begin{tabular}{ccccc}
    \toprule
    Benchmark & 
    \multicolumn{2}{c}{Pitzdaily} & \multicolumn{2}{c}{ESP}\\
    \cmidrule(r){1-1}\cmidrule(lr){2-3}\cmidrule(lr){4-5}
    Model & 
    Train & Test & Train & Test\\
    \midrule
    PWL on IDONE & 
    - & - & 
    $2\cdot 10^{-4}$ & $11.16$ \\
    PWL on MVRSM &
    $0.047$ & $4.092$ & 
    $0.002$ & $11.58$ \\
    RF on SMAC &
    $0.101$ & $1.151$ &
    $0.148$ & $0.855$ \\
    GP on BO & 
    $9\cdot 10^{-7}$ & $\mathbf{0.915}$ &
    $5.5\cdot 10^{-7}$ & $\mathbf{0.835}$ \\
    GP on DONE & 
    $\mathbf{5\cdot 10^{-8}}$ & $1.888$ &
    $\mathbf{4.8\cdot 10^{-7}}$ & $0.910$ \\
    \bottomrule
    \end{tabular}
\end{table}

\subsection{\editB{Summary of obtained insights}}\label{sec:Disc}

Based on the \editB{experimental results}, 
we highlight the most important insights that were obtained.
First of all, the \emph{type of variable} a surrogate model is designed for, is not necessarily a good indicator of the performance of the surrogate algorithm in case of a real-life problem: discrete surrogates can perform well on continuous problems, and vice versa.
We saw this on the wind farm layout optimisation problem, a continuous problem where a discrete surrogate model (SMAC's random forest) had the best performance, and on the ESP problem, a discrete problem where the continuous Gaussian process surrogate model had the best performance even though it was unable to outperform random search on the wind farm layout optimisation problem.
\editA{For the latter problem, changing the kernel type or parameters of the Gaussian process might improve results, however we expected the chosen settings to work well for the problem.}
Part of these insights were known from previous work~\cite{BNAIC2020paper}, but we extended these insights to continuous problems and to more benchmark problems and surrogate algorithms.
\editB{
The claim is supported by rules of thumb 4 and 6 of Section~\ref{sec:thumb}, while the other rules of thumb were more in line with expectations.}
The experiments using offline learning of machine learning models also showed that discrete models such as random forests can have lower generalisation error than continuous models, even on data coming from a continuous problem like Pitzdaily.
\editB{This result is surprising, considering random forests are known to have poor extrapolation capabilities.}

Second, our observations lead us to believe that \emph{exploration is more important than model accuracy} in surrogate algorithms.
The offline learning experiments  showed that surrogate models trained on data gathered by an algorithm that uses that model are not necessarily more accurate than surrogate models trained on data gathered by random search, a high-exploration method.
The use of random search should also not be underestimated, as the experiment where we artificially change the evaluation time of the objective shows that for all considered benchmarks there are situations where random search outperforms all surrogate algorithms, mainly when the objective evaluation time is low.
\editB{This is supported by the first three rules of thumb in Section~\ref{sec:thumb}.}
Furthermore, on the ESP problem, MVRSM had a much better performance than IDONE, even though they use exactly the same piece-wise linear surrogate model on that problem.
\editB{For discrete problems,} the only difference between the two algorithms is that MVRSM has a higher exploration rate.
The low training error of the piece-wise linear surrogate model shows that \editA{for the considered problems,} a highly accurate model does not necessarily lead to a better performance of the surrogate algorithm using that model.

Finally, \emph{the available time budget and the evaluation time of the objective} strongly influence which algorithm is the best choice for a certain problem.
This can be seen from the experiment where we artificially change the function evaluation time:  the best performing algorithm then varies depending on the available time budget and function evaluation time.
\editB{This claim is supported by all rules of thumb in Section~\ref{sec:thumb}. In fact, the evaluation time of the objective was the most important feature of the decision tree classifier that we used to generate the rules of thumb.}

\section{Conclusion and future work}\label{sec:concl}


We proposed a public benchmark library called \bname, which
fills an important gap in the current landscape of optimisation benchmark libraries that mostly consists of cheap to evaluate benchmark functions or of expensive problems with no or limited baseline solutions from surrogate model literature.
\editB{This resulted in a dataset containing the results of running multiple surrogate-based optimisation algorithms on several expensive problems, which can be used to create tabular or surrogate benchmarks or for meta-learning.}
A first analysis of this dataset 
showed how the best choice of algorithm for a certain problem depends on the available time budget and the evaluation time of the objective, and we provided a method to extrapolate such results to real-life problems that contain expensive objective functions with different costs.
\editB{We also provided easy to interpret rules of thumb based on our analysis, showing when to use which surrogate algorithm.
Furthermore, }the dataset allowed us to train surrogate models offline rather than online, giving insight into the generalisation capabilities of the surrogate models and showing the potential of models such as XGBoost to be used in new surrogate algorithms in the future.
Finally, we showed how continuous models can work well for discrete problems and vice versa, and we highlighted the important role of exploration in surrogate algorithms.
In future work we will focus on methods that can deal with the constraints present in some of the benchmark problems from this work, as well as make a comparison with surrogate-assisted evolutionary methods%
\editA{, particularly for multi-objective problems}.

\section*{Acknowledgments}
This work is part of the research programme Real-time data-driven maintenance logistics with project number 628.009.012, which is financed by the Dutch Research Council (NWO).

\appendix

\section{Area under curve metrics}\label{sec:aoc}

This section shows the area under the curve (AUC) of the best found objective value for each method on each problem in the benchmark library. 
See Table~\ref{tab:AUC}.
Note that here we used a different normalisation procedure before calculating the AUC: the random initial samples were included, and then for every problem, results were normalised as follows:
\begin{align}
    f_{norm} = (f-f_{max})/(f_{min}-f_{max}),
\end{align}
where $f_{max}$ ($f_{min}$) is the best (worst) found objective function across all methods and iterations for a particular problem.

\begin{table*}[htb]
\setlength{\tabcolsep}{3pt}
    \centering
    \caption{Area under the curve metrics after $500$ and $1000$ iterations (higher is better), averaged over multiple runs.}
    \label{tab:AUC}
    \resizebox{\textwidth}{!}{
    \begin{tabular}{rcccccccc}
    \toprule
    Benchmark & 
    \multicolumn{2}{c}{Windwake} & 
    \multicolumn{2}{c}{Pitzdaily} & \multicolumn{2}{c}{ESP} & \multicolumn{2}{c}{HPO} \\
    \cmidrule(r){1-1}\cmidrule(lr){2-3}\cmidrule(lr){4-5}\cmidrule(lr){6-7}\cmidrule(lr){8-9}
    Algorithm\\ at iteration& 
    $500$ & $1000$ &
    $500$ & $1000$ & $500$ & $1000$ & $500$ & $1000$\\
    \midrule
    RS 	& $0.963844$ &  $0.968033$	 & 	$0.976079$ & $0.983226$	 & 	$0.773124$ & $0.804465$	 & 	$0.930773$ & $0.935801$	\\
    BO 	& $0.965312$ &  $0.969517$	 & 	$0.986475$ & $0.989424$	 & 	$\mathbf{0.872638}$ & $\mathbf{0.909874}$	 & 	$0.925708$ & $0.936327$	\\
    HO 	& $0.970493$ &  $0.975888$	 & 	$0.986173$ & $0.990593$	 & 	$0.810760$ & $0.849704$	 & 	$\mathbf{0.937372}$ & $\mathbf{0.952699}$	\\
    SMAC 	& $\mathbf{0.980197}$ &  $\mathbf{0.984984}$	 & 	$\mathbf{0.989619}$ & $\mathbf{0.993185}$	 & 	$0.827741$ & $0.870288$	 & 	$0.932997$ & $0.951097$	\\
    DONE 	& $0.974548$ &  $0.979858$	 & 	$0.953432$ & $0.955938$	 & 	$0.840280$ & $0.869861$	 & 	$0.923643$ & $0.935636$	\\
    IDONE 	& - &  -	 & -	 & -	 & 	$0.787532$ & $0.812625$	 & - 	 & -	\\
    MVRSM 	& $0.976244$ &  $0.980730$	 & 	$0.987027$ & $0.991660$	 & 	$0.850981$ & $0.887566$	 & 	$0.933505$ & $0.945031$	\\
    \bottomrule
    \end{tabular}}
\end{table*}

\section{Offline learning results}\label{app:offline}

Here we show more data of the offline learning experiment presented in Section~IV.D%
, including the standard deviations.
See Table~\ref{tab:offlineAll}.
No results are given for the hyperparameter optimisation benchmark, as not all supervised learning models are able to deal with the conditional variables present in this benchmark.
New models that were not explained in the main text are two models from scikit-learn: a polynomial model with degree $2$ (Quadratic), and a multi-layer perceptron with default hyperparameters (MLP).

\begin{table*}[htbp]
\setlength{\tabcolsep}{3pt}
    \centering
    \caption{Mean average error for models trained on data gathered by one surrogate algorithm, averaged over different runs. }
    \label{tab:offlineAll}
    \resizebox{\textwidth}{!}{
    \begin{tabular}{rcccccc}
    \toprule
    Benchmark & 
    \multicolumn{2}{c}{Windwake} & 
    \multicolumn{2}{c}{Pitzdaily} & \multicolumn{2}{c}{ESP}\\
    \cmidrule(r){1-1}\cmidrule(lr){2-3}\cmidrule(lr){4-5}\cmidrule(lr){6-7}
    Model$+$method & 
    Train & Test &
    Train & Test & Train & Test\\
    \midrule
    Linear on RS & $3.3\cdot 10^{10} \pm 3\cdot 10^8$ & $3.7\cdot 10^{10} \pm 3\cdot 10^9$     &
    $0.70\pm 0.03$ & $1.23 \pm 0.15$ & 
    $2.7 \pm 6.0$ & $1.8 \pm 2.3$ \\
    Quadratic on RS & $2.8\cdot 10^{10} \pm 6\cdot 10^8$ & $\mathbf{1.9\cdot 10^{10}} \pm 5\cdot 10^9$     &
    $0.50\pm 0.01$ & $1.22 \pm 0.13$ & 
    $\mathbf{3\cdot 10^{-14}} \pm 8\cdot 10^{-14}$ & $4.4 \pm 8.8$ \\
    PWL on RS & $3.1\cdot 10^{10} \pm 1\cdot 10^9$ & $2.0\cdot 10^{10} \pm 4\cdot 10^9$     &
    $6\cdot 10^{-3}\pm 6\cdot 10^{-4}$ & $1.32 \pm 0.44$ & 
    $5 \cdot 10^{-9} \pm 1\cdot 10^{-8}$ & $8.9 \pm 16.9$ \\
    RF on RS & $1.2\cdot 10^{10} \pm 2\cdot 10^8$ & $3.5\cdot 10^{10} \pm 2\cdot 10^9$     &
    $0.15\pm 0.01$ & $\mathbf{0.76} \pm 0.17$ & 
    $0.6 \pm 1.2$ & $1.0 \pm 0.4$ \\
    XGBoost on RS & $1.6\cdot 10^{10} \pm 1\cdot 10^9$ & $3.7\cdot 10^{10} \pm 4\cdot 10^9$     &
    $0.28\pm 0.01$ & $0.85 \pm 0.23$ & 
    $0.2 \pm 0.2$ & $\mathbf{0.7} \pm 0.1$ \\
    GP on RS & $\mathbf{3\cdot 10^{4}} \pm 1\cdot 10^2$ & $3.7\cdot 10^{10} \pm 2\cdot 10^9$     &
    $\mathbf{8\cdot 10^{-7}}\pm 3\cdot 10^{-8}$ & $1.15 \pm 0.09$ & 
    $1\cdot 10^{-6}\pm 2\cdot 10^{-6}$ & $1.3 \pm 1.0$ \\
    MLP on RS & $3.5\cdot 10^{10} \pm 2\cdot 10^9$ & $7.3\cdot 10^{10} \pm 2\cdot 10^2$     &
    $0.73\pm 0.03$ & $1.36 \pm 0.02$ & 
    $0.4\pm 0.9$ & $0.9 \pm 0.6$ \\
    \midrule
    Linear on BO & $3.2\cdot 10^{10} \pm 2\cdot 10^9$ & $3.7\cdot 10^{10} \pm 5\cdot 10^9$     &
    $0.77\pm 0.04$ & $0.95 \pm 0.35$ & 
    $0.2 \pm 5 \cdot 10^{-2}$ & $1.2 \pm 0.2$ \\
    Quadratic on BO & $2.6\cdot 10^{10} \pm 2\cdot 10^9$ & $2.4\cdot 10^{10} \pm 1.0\cdot 10^{10}$     &
    $0.53\pm 0.15$ & $1.34 \pm 0.21$ & 
    $\mathbf{5\cdot 10^{-15}} \pm 3\cdot 10^{-15}$ & $2.0 \pm 2.0$ \\
    PWL on BO & $2.9\cdot 10^{10} \pm 2\cdot 10^9$ & $\mathbf{2.0\cdot 10^{10}} \pm 3\cdot 10^9$     &
    $0.01\pm 2\cdot 10^{-3}$ & $1.80 \pm 0.93$ & 
    $1 \cdot 10^{-9} \pm 7\cdot 10^{-10}$ & $4.0 \pm 2.1$ \\
    RF on BO & $1.1\cdot 10^{10} \pm 1\cdot 10^9$ & $3.7\cdot 10^{10} \pm 3\cdot 10^9$     &
    $0.15\pm 0.01$ & $1.05 \pm 0.15$ & 
    $7\cdot 10^{-2} \pm 2\cdot 10^{-2}$ & $1.0 \pm 0.3$ \\
    XGBoost on BO & $1.5\cdot 10^{10} \pm 2\cdot 10^9$ & $4.1\cdot 10^{10} \pm 9\cdot 10^9$     &
    $0.24\pm 0.02$ & $1.00 \pm 0.28$ & 
    $8 \cdot 10^{-2} \pm 7\cdot 10^{-3}$ & $\mathbf{0.7} \pm 9 \cdot 10^{-2}$ \\
    GP on BO & $\mathbf{3\cdot 10^{4}} \pm 1\cdot 10^3$ & $3.4\cdot 10^{10} \pm 5\cdot 10^9$     &
    $\mathbf{9\cdot 10^{-7}}\pm 2\cdot 10^{-8}$ & $\mathbf{0.92} \pm 0.05$ & 
    $5\cdot 10^{-7}\pm 3\cdot 10^{-7}$ & $0.8 \pm 0.2$ \\
    MLP on BO & $3.9\cdot 10^{10} \pm 5\cdot 10^9$ & $7.3\cdot 10^{10} \pm 1\cdot 10^2$     &
    $0.68\pm 0.04$ & $1.10 \pm 0.28$ & 
    $0.1\pm 5\cdot 10^{-2}$ & $1.3 \pm 0.5$ \\
    \midrule
    Linear on HO & $2.6\cdot 10^{10} \pm 2\cdot 10^9$ & $4.6\cdot 10^{10} \pm 5\cdot 10^9$     &
    $0.77\pm 0.04$ & $0.95 \pm 0.35$ & 
    $0.2 \pm 0.1$ & $0.8 \pm 0.2$ \\
    Quadratic on HO & $2.2\cdot 10^{10} \pm 2\cdot 10^9$ & $3.7\cdot 10^{10} \pm 9\cdot 10^9$     &
    $0.53\pm 0.15$ & $1.34 \pm 0.21$ & 
    $2\cdot 10^{-3} \pm 2\cdot 10^{-3}$ & $3\cdot 10^9 \pm 6\cdot 10^9$ \\
    PWL on HO & $2.5\cdot 10^{10} \pm 2\cdot 10^9$ & $2.9\cdot 10^{10} \pm 9\cdot 10^9$     &
    $0.01\pm 2\cdot 10^{-3}$ & $1.80 \pm 0.93$ & 
    $3 \cdot 10^{-5} \pm 6\cdot 10^{-5}$ & $4.3 \pm 3.4$ \\
    RF on HO & $1.0\cdot 10^{10} \pm 8\cdot 10^8$ & $3.6\cdot 10^{10} \pm 2\cdot 10^9$     &
    $0.15\pm 0.01$ & $1.05 \pm 0.15$ & 
    $6 \cdot 10^{-2} \pm 3\cdot 10^{-2}$ & $0.7 \pm 5\cdot 10^{-2}$ \\
    XGBoost on HO & $1.1\cdot 10^{10} \pm 2\cdot 10^9$ & $4.3\cdot 10^{10} \pm 5\cdot 10^9$     &
    $0.24\pm 0.02$ & $1.00 \pm 0.28$ & 
    $7\cdot 10^{-2} \pm 10^{-2}$ & $\mathbf{0.6} \pm 8\cdot 10^{-2}$ \\
    GP on HO & $\mathbf{3\cdot 10^{4}} \pm 3\cdot 10^3$ & $\mathbf{2.5\cdot 10^{10}} \pm 2\cdot 10^9$     &
    $\mathbf{9\cdot 10^{-7}}\pm 2\cdot 10^{-8}$ & $\mathbf{0.92} \pm 0.05$ & 
    $\mathbf{3\cdot 10^{-7}}\pm 2\cdot 10^{-7}$ & $0.7 \pm 0.1$ \\
    MLP on HO & $4.8\cdot 10^{10} \pm 3\cdot 10^9$ & $7.3\cdot 10^{10} \pm 1\cdot 10^2$     &
    $0.68\pm 0.04$ & $1.10 \pm 0.28$ & 
    $6\cdot 10^{-2}\pm 10^{-2}$ & $0.7 \pm 0.1$ \\
    \midrule
    Linear on SMAC & $1.2\cdot 10^{10} \pm 1\cdot 10^9$ & $3.0\cdot 10^{10} \pm 4\cdot 10^9$     &
    $0.47\pm 0.10$ & $0.83 \pm 0.48$ & 
    $0.4 \pm 0.3$ & $1.0 \pm 0.4$ \\
    Quadratic on SMAC & $9\cdot 10^{9} \pm 1\cdot 10^9$ & $5.0\cdot 10^{10} \pm 1.8\cdot 10^{10}$     &
    $0.39\pm 0.08$ & $1.24 \pm 0.82$ & 
    $\mathbf{6\cdot 10^{-15}} \pm 5\cdot 10^{-15}$ & $0.9 \pm 0.4$ \\
    PWL on SMAC & $1.2\cdot 10^{10} \pm 1\cdot 10^9$ & $1.8\cdot 10^{10} \pm 4\cdot 10^9$     &
    $0.09\pm 0.03$ & $2.91 \pm 0.91$ & 
    $1 \cdot 10^{-9} \pm 5\cdot 10^{-10}$ & $2.5 \pm 1.3$ \\
    RF on SMAC & $4\cdot 10^{9} \pm 4\cdot 10^8$ & $3.4\cdot 10^{10} \pm 3\cdot 10^9$     &
    $0.10\pm 0.02$ & $1.15 \pm 0.27$ & 
    $0.1 \pm 0.1$ & $0.9 \pm 0.1$ \\
    XGBoost on SMAC & $4\cdot 10^{9} \pm 5\cdot 10^8$ & $4.4\cdot 10^{10} \pm 8\cdot 10^9$     &
    $0.11\pm 0.03$ & $1.11 \pm 0.31$ & 
    $0.1 \pm 3\cdot 10^{-2}$ & $\mathbf{0.7} \pm 6 \cdot 10^{-2}$ \\
    GP on SMAC & $\mathbf{8\cdot 10^{4}} \pm 5\cdot 10^4$ & $\mathbf{1.3\cdot 10^{10}} \pm 1\cdot 10^9$     &
    $\mathbf{9\cdot 10^{-7}}\pm 3\cdot 10^{-7}$ & $\mathbf{0.44} \pm 0.09$ & 
    $3\cdot 10^{-7}\pm 2\cdot 10^{-7}$ & $0.8 \pm 0.1$ \\
    MLP on SMAC & $6.0\cdot 10^{10} \pm 1\cdot 10^9$ & $7.3\cdot 10^{10} \pm 2\cdot 10^2$     &
    $0.46\pm 0.08$ & $0.83 \pm 0.45$ & 
    $6 \cdot 10^{-2}\pm 3 \cdot 10^{-2}$ & $0.9 \pm 0.4$ \\
    \midrule
    Linear on DONE & $2.6\cdot 10^{10} \pm 10^9$ & $5.5\cdot 10^{10} \pm 3\cdot 10^9$     &
    $0.04\pm 0.01$ & $1.90 \pm 0.008$ & 
    $0.2 \pm 2\cdot 10^{-2}$ & $0.9 \pm 0.1$ \\
    Quadratic on DONE & $2.3\cdot 10^{10} \pm 10^9$ & $5.8\cdot 10^{10} \pm 2\cdot 10^{9}$     &
    $0.08\pm 0.02$ & $\mathbf{1.55} \pm 0.13$ & 
    $\mathbf{4\cdot 10^{-15}} \pm 2\cdot 10^{-15}$ & $1.0 \pm 0.2$ \\
    PWL on DONE & $2.5\cdot 10^{10} \pm 10^9$ & $5.5\cdot 10^{10} \pm 2\cdot 10^9$     &
    $6\cdot 10^{-4}\pm 10^{-4}$ & $1.78 \pm 0.10$ & 
    $8 \cdot 10^{-10} \pm 9\cdot 10^{-11}$ & $2.1 \pm 1.0$ \\
    RF on DONE & $9\cdot 10^{9} \pm 4\cdot 10^8$ & $\mathbf{5.2\cdot 10^{10}} \pm 3\cdot 10^9$     &
    $0.01\pm 3\cdot 10^{-3}$ & $1.70 \pm 0.09$ & 
    $5\cdot 10^{-2} \pm 9\cdot 10^{-3}$ & $0.8 \pm 6\cdot 10^{-2}$ \\
    XGBoost on DONE & $1.0\cdot 10^{10} \pm 8\cdot 10^8$ & $5.4\cdot 10^{10} \pm 5\cdot 10^9$     &
    $0.05\pm 2\cdot 10^{-3}$ & $1.60 \pm 0.14$ & 
    $6\cdot 10^{-2} \pm 4\cdot 10^{-3}$ & $\mathbf{0.7} \pm 6\cdot 10^{-2}$ \\
    GP on DONE & $\mathbf{3\cdot 10^{4}} \pm 10^3$ & $5.5\cdot 10^{10} \pm 2\cdot 10^9$     &
    $\mathbf{5\cdot 10^{-8}}\pm 2\cdot 10^{-8}$ & $1.89 \pm 0.07$ & 
    $5\cdot 10^{-7}\pm 2\cdot 10^{-7}$ & $0.9 \pm 4\cdot 10^{-2}$ \\
    MLP on DONE & $1.8\cdot 10^{10} \pm 10^9$ & $7.3\cdot 10^{10} \pm 2\cdot 10^2$     &
    $0.08\pm 5\cdot 10^{-3}$ & $1.91 \pm 0.009$ & 
    $0.1\pm 7\cdot 10^{-2}$ & $0.8 \pm 0.2$ \\
    \midrule
    Linear on IDONE & - & -     &
    - & - & 
    $0.2 \pm 3\cdot 10^{-2}$ & $1.0 \pm 0.2$ \\
    Quadratic on IDONE & - & -     &
    - & - & 
    $\mathbf{4\cdot 10^{-15}} \pm 3\cdot 10^{-15}$ & $1.6 \pm 1.3$ \\
    PWL on IDONE & - & -     &
    - & - & 
    $2\cdot 10^{-4} \pm 3\cdot 10^{-4}$ & $11.2 \pm 2.6$ \\
    RF on IDONE & - & -     &
    - & - & 
    $4\cdot 10^{-2} \pm 9\cdot 10^{-3}$ & $1.2 \pm 0.5$ \\
    XGBoost on IDONE & - & -     &
    - & - & 
    $6\cdot 10^{-2} \pm 10^{-2}$ & $\mathbf{0.8} \pm 0.1$ \\
    GP on IDONE & - & -     &
    - & - & 
    $6\cdot 10^{-7} \pm 4\cdot 10^{-7}$ & $0.9 \pm 0.1$ \\
    MLP on IDONE & - & -     &
    - & - & 
    $8\cdot 10^{-2} \pm 4\cdot 10^{-2}$ & $1.0 \pm 0.4$ \\
    \midrule
    Linear on MVRSM & $7\cdot 10^{9} \pm 2\cdot 10^9$ & $4.5\cdot 10^{10} \pm 1.1\cdot 10^{10}$     &
    $0.48\pm 0.07$ & $1.16\pm 0.35$ & 
    $0.1 \pm 3\cdot 10^{-2}$ & $0.8 \pm 0.2$ \\
    Quadratic on MVRSM & $8\cdot 10^{9} \pm 4\cdot 10^9$ & $1.75\cdot 10^{11} \pm 6.1\cdot 10^{10}$     &
    $0.40\pm 0.15$ & $0.96 \pm 0.96$ & 
    $2\cdot 10^{-4} \pm 3\cdot 10^{-4}$ & $5\cdot 10^{7} \pm 5\cdot 10^{7}$ \\
    PWL on MVRSM & $7\cdot 10^{9} \pm 2\cdot 10^9$ & $5.2\cdot 10^{10} \pm 1.3\cdot 10^{10}$     &
    $0.05\pm 0.02$ & $4.09 \pm 1.90$ & 
    $2 \cdot 10^{-3} \pm 10^{-3}$ & $11.6 \pm 2.8$ \\
    RF on MVRSM & $2\cdot 10^{9} \pm 5\cdot 10^8$ & $4.4\cdot 10^{10} \pm 6\cdot 10^9$     &
    $0.06\pm 10^{-3}$ & $1.21 \pm 0.29$ & 
    $3\cdot 10^{-2} \pm 7\cdot 10^{-3}$ & $0.8 \pm 0.2$ \\
    XGBoost on MVRSM & $2\cdot 10^{9} \pm 2\cdot 10^8$ & $5.1\cdot 10^{10} \pm 1.0\cdot 10^{10}$     &
    $0.06\pm 7\cdot 10^{-3}$ & $1.16 \pm 0.40$ & 
    $4\cdot 10^{-2} \pm 10^{-2}$ & $0.7 \pm 0.3$ \\
    GP on MVRSM & $\mathbf{2\cdot 10^{5}} \pm 2\cdot 10^5$ & $\mathbf{1.3\cdot 10^{10}} \pm 6\cdot 10^9$     &
    $\mathbf{8\cdot 10^{-7}}\pm 2\cdot 10^{-7}$ & $\mathbf{0.53} \pm 0.07$ & 
    $\mathbf{5\cdot 10^{-7}}\pm 3\cdot 10^{-7}$ & $\mathbf{0.5} \pm 0.2$ \\
    MLP on MVRSM & $6.4\cdot 10^{10} \pm 3\cdot 10^9$ & $7.3\cdot 10^{10} \pm 1\cdot 10^2$     &
    $0.37\pm 0.09$ & $1.18 \pm 0.34$ & 
    $9\cdot 10^{-2}\pm 3\cdot 10^{-2}$ & $0.7 \pm 0.3$ \\
    \bottomrule
    \end{tabular}}
\end{table*}

\bibliographystyle{elsarticle-num}
\bibliography{docbib,challengebib}

\end{document}